%% file: root.tex
\documentclass[letterpaper, 10 pt, conference]{ieeeconf}  

\usepackage{algorithm}
\usepackage{algpseudocode}
\usepackage{amsmath}

\IEEEoverridecommandlockouts                              

\input{definitions}

\title{\LARGE \bf \textit{OVAMOS}: A Framework for Open-Vocabulary Multi-Object Search in Unknown Environments}

\author{Qianwei Wang*, Yifan Xu*, Vineet Kamat, and Carol Menassa
\thanks{* Main Contribution}
\thanks{Funding for V. Kamat, C. Menassa and Y. Xu was provided by NSF Award No. 2124857.}
\thanks{The authors are with the University of Michigan, Ann Arbor, MI 48109, USA. {\tt\small\{qweiw, yfx, vkamat, menassa\}@umich.edu}}%
}

\begin{document}

\maketitle
\thispagestyle{empty}
\pagestyle{empty}

\begin{abstract}
Object search is a fundamental task for robots deployed in indoor building environments, yet challenges arise due to observation instability, especially for open-vocabulary models. While foundation models (LLMs/VLMs) enable reasoning about object locations even without direct visibility, the ability to recover from failures and replan remains crucial. The Multi-Object Search (MOS) problem further increases complexity, requiring the tracking multiple objects and thorough exploration in novel environments, making observation uncertainty a significant obstacle. To address these challenges, we propose a framework integrating VLM-based reasoning, frontier-based exploration, and a Partially Observable Markov Decision Process (POMDP) framework to solve the MOS problem in novel environments. VLM enhances search efficiency by inferring object-environment relationships, frontier-based exploration guides navigation in unknown spaces, and POMDP models observation uncertainty, allowing recovery from failures in occlusion and cluttered environments. We evaluate our framework on 120 simulated scenarios across several Habitat-Matterport3D (HM3D) scenes and a real-world robot experiment in a 50-square-meter office, demonstrating significant improvements in both efficiency and success rate over baseline methods.

\end{abstract}

\IEEEpeerreviewmaketitle

\input{intro}

\input{relatedwork}

\input{Methodology}

\input{results}

\input{conclusion}

{\small 
\balance
\bibliographystyle{IEEEtran}
\bibliography{bib/strings-abrv,bib/ieee-abrv,bib/references}
}

\end{document}

%% file: definitions.tex
\usepackage{graphicx}
\usepackage{caption}
\usepackage{caption}
\usepackage{subcaption}
\usepackage{amsmath,amssymb,enumerate}

\usepackage{amsthm}
\usepackage{setspace}
\usepackage{amsmath,amssymb,enumerate}
\usepackage{booktabs}
\usepackage[usenames,dvipsnames,svgnames,table]{xcolor}
\usepackage{mathtools}
\usepackage{algorithm, algorithmicx, algpseudocode}

\usepackage{blindtext}
\usepackage{gensymb}
\usepackage{xparse}
\usepackage{lipsum}
\usepackage{mathrsfs}
\usepackage[mathscr]{euscript}
\usepackage{times}
\usepackage[noadjust]{cite} 
\usepackage{multicol}
\usepackage{multirow}
\definecolor{darkgreen}{rgb}{0,0.6,0}
\usepackage[bookmarks=true,colorlinks=true,pdfpagemode=UseNone,citecolor=black,linkcolor=black,urlcolor=BrickRed]{hyperref}
\usepackage{amsfonts}
\usepackage{cleveref}
\usepackage[utf8]{inputenc}
\usepackage[T1]{fontenc}
\usepackage{textcomp}
\usepackage{arydshln}
\usepackage{tabu}
\usepackage{cuted}
\usepackage{balance}
\usepackage{svg}
\usepackage{graphicx}
\usepackage{makecell}
\usepackage{booktabs}
\usepackage{multirow}
\usepackage{bbding}

\definecolor{note}{rgb}{0.1,0.1,1}
\definecolor{rephase}{rgb}{0.15,0.7,0.15}
\definecolor{bag}{rgb}{0.6,0.6,0.2}

\makeatletter
\renewcommand*\env@matrix[1][c]{\hskip -\arraycolsep
  \let\@ifnextchar\new@ifnextchar
  \array{*\c@MaxMatrixCols #1}}
\makeatother

\DeclareDocumentCommand{\vector}{ O{} }{\mathrm{vec}(#1)}

\makeatletter
\newcommand{\mathleft}{\@fleqntrue\@mathmargin0pt}
\newcommand{\mathcenter}{\@fleqnfalse}
\makeatother

%% file: intro.tex
\section{Introduction}
\label{sec:intro}

Multi-Object Search (MOS) is a crucial task in robotics~\cite{gen}. Consider a scenario where in a workplace setting, a robot may need to retrieve multiple objects to complete a task, such as gathering necessary documents, tools, or equipment for an assembly process. Similarly, in household environments, a robot making a coffee or preparing a hamburger must locate various ingredients and kitchen tools before assembling the final products.  

\begin{figure}[t]
     \centering
     \includegraphics[width=\columnwidth]{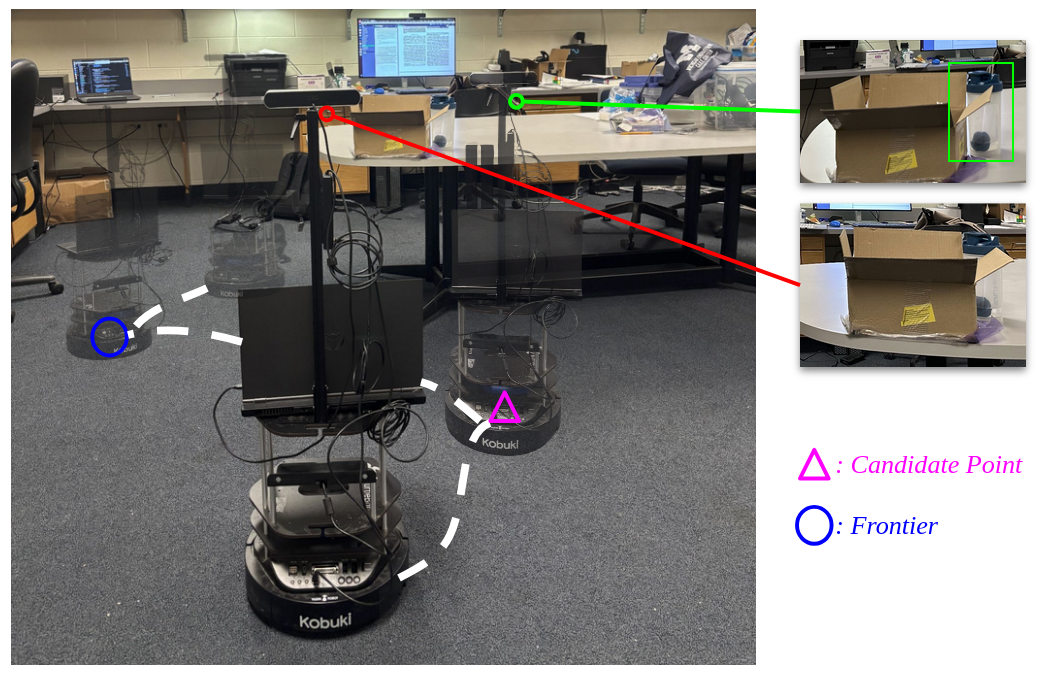}
    \caption{A TurtleBot deployed with OVAMOS searching for multiple objects in a cluttered office environment. Initially, the detector fails to recognize a partially occluded water bottle. Instead of immediately exploring new frontiers, OVAMOS leverages VLM-guided information to prioritize candidate points, gathering additional viewpoints in high-value regions. This strategy ultimately enables the successful detection of the target object.}
    \label{fig: intro}
\end{figure}

In object search problems, one significant challenge is the uncertainty in observations~\cite{lessons}. For instance, as shown in Fig.~\ref{fig: intro}, when a cardboard box partially occludes a water cup, the detector may fail to recognize it. Error detection and recovery become especially crucial as perception reliability decreases with environmental complexity. Consequently, in MOS tasks—where multiple objects must be tracked, and the environment explored in depth—it is vital that agents continuously plan and reason about alternative actions to gather additional viewpoints and compensate for any missed detections.

Existing approaches to MOS can be broadly categorized into probabilistic planning, Deep Reinforcement Learning (DRL), and foundation model--based methods. Probabilistic planning methods~\cite{mos_base,mos_system,mos_new}, often formulated as Partially Observable Markov Decision Processes (POMDPs), manage uncertainty in object locations and perception by maintaining belief states and planning under partial observability. While effective for tracking multiple objects and enabling thorough exploration, these methods suffer from significant computational inefficiencies when scaling to large environments due to the complexity of belief state updates over extended planning horizons. DRL-based approaches~\cite{multion,learning_long,multi_drl,multi_drl_2}, in contrast, train robots through repeated interactions with the environment to develop exploration strategies. Although capable of learning effective search policies, they often struggle with inefficient exploration and poor generalizability, particularly in novel environments. More recently, foundation model--based approaches, which encompass  Large Language Models (LLMs)~\cite{esc,zson,cat,lvmn} or Vision-Language models (VLMs)~\cite{opennav,vlfm}, have improved Single-Object Search (SOS) by leveraging object--scene and object--object relationships for more efficient target localization. However, existing multi-object search frameworks~\cite{finder} incorporating foundation models often fail to address the challenges of observation uncertainty and complex environments, limiting their ability to efficiently navigate complex, partially observed spaces.

To address these limitations, we propose a novel framework that integrates VLMs, frontier-based exploration, and POMDP-based planning. Our approach utilizes VLMs to construct a multi-layer value map for different target objects. Instead of treating this map as static, we introduce a Bayesian-inspired decay mechanism, where the value of regions decreases over time if the detector repeatedly fails to locate objects. We then apply DBSCAN~\cite{dbscan} clustering to transform the value map into belief representations and candidate points for POMDP-based action selection, effectively incorporating VLM-derived information into the POMDP formulation. Finally, we incorporate reward design for frontier-based exploration and solve the POMDP using Partially Observable Upper Confidence Trees (POUCT)~\cite{pouct}.

Our main contributions are as follows:
\begin{itemize}
    \item \textbf{A unified multi-object search framework:} We propose OVAMOS, an open-vocabulary multi-object search framework that integrates VLMs, frontier-based exploration, and POMDP-based planning to achieve efficient and robust multi-object search.
    \item \textbf{VLM-guided POMDP:} We leverage VLMs to simplify POMDP belief updates and action selection, enabling effective uncertainty handling while mitigating the computational challenges of solving large-scale POMDPs.
    \item \textbf{Extensive evaluation and real-world validation:} We validate OVAMOS in 120 simulated episodes across 5 HM3D~\cite{hm3d} scenes and conduct real-world experiments in a 50 m$^2$ office, demonstrating superior efficiency and success rates compared to baselines.
\end{itemize}

\begin{figure*}[t!]
    \centering
    \includegraphics[width = \textwidth]{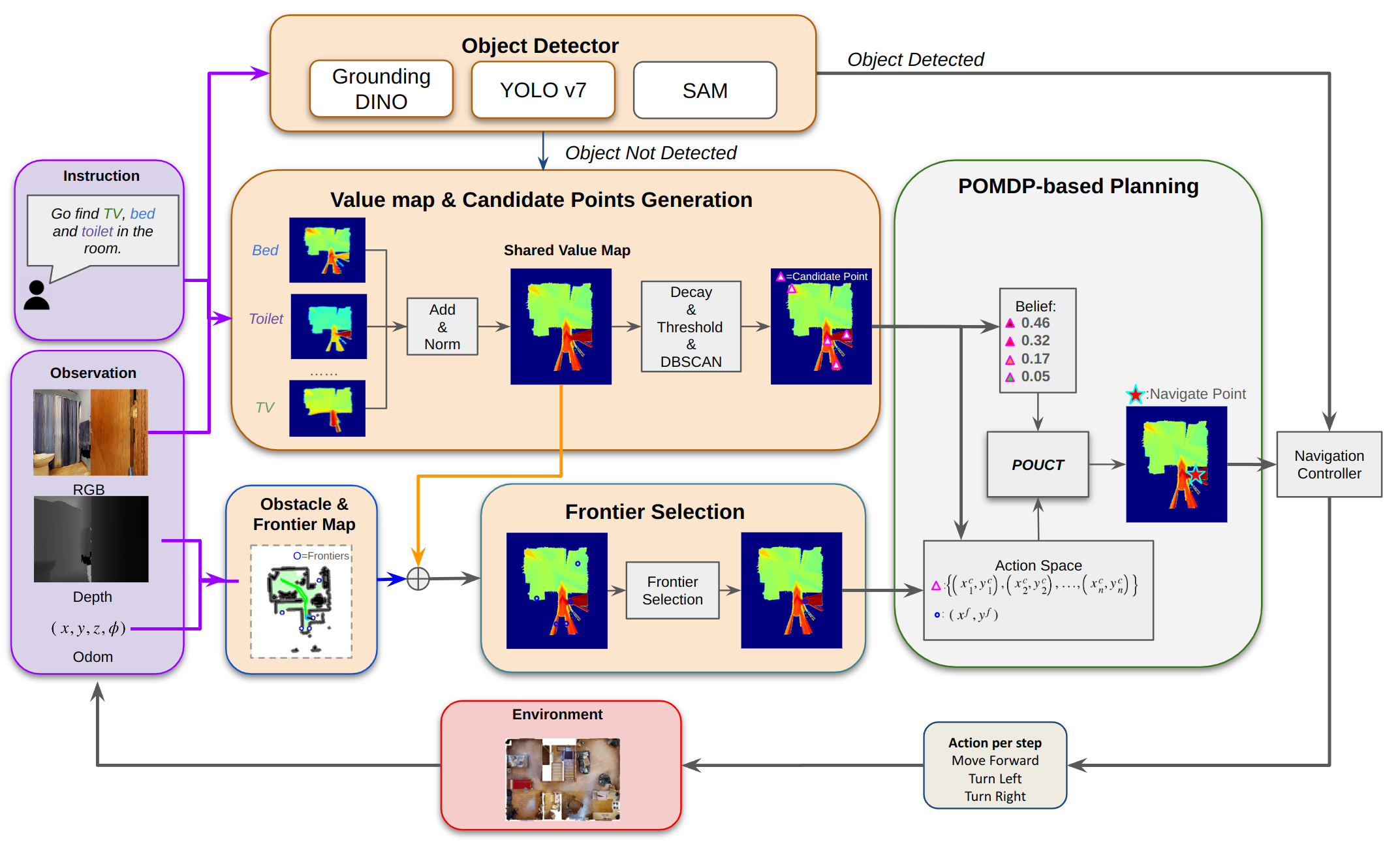}
    \caption{\textbf{OVAMOS} consists of a \textit{mapping module}, a \textit{planning module}, and a \textit{navigation controller}. The mapping module processes RGB-D inputs and textual prompts to generate an \textit{object-value map}, which integrates detected objects and estimated potential object locations. If a target object is found in the object map, the robot navigates directly to it; otherwise, it relies on the value map to guide exploration. Additionally, the module constructs an obstacle map for navigation constraints and a frontier map to identify unexplored areas. The planning module maintains a belief representation of object locations and employs the POUCT algorithm to simulate and evaluate action sequences, selecting the one with the highest expected reward for execution. The \textit{navigation controller} receives a target location from the planning module and outputs discrete movement commands (\textit{move forward, turn left, turn right}) to guide the robot toward its destination.}
    \label{fig:system_structure}
\end{figure*}

%% file: relatedwork.tex
\section{Related Work}
\label{sec:relatedwork}

Existing approaches to MOS can be broadly categorized into three types: DRL methods~\cite{multion,learning_long,multi_drl,multi_drl_2}, probabilistic planning methods~\cite{mos_base,mos_new,mos_system}, and foundation model--guided navigation methods. Most prior work falls into the first two categories, while foundation model--guided approaches—which leverage LLMs~\cite{esc,cat,lvmn,zson} or VLMs~\cite{vlfm,opennav}—were initially popular in SOS, they have recently been extended to MOS~\cite{finder}.

\subsection{Deep Reinforcement Learning Methods for MOS}
DRL-based methods, including Deep Q Networks (DQN)~\cite{dqn}, Proximal Policy Optimization (PPO)~\cite{multion,learning_long,multi_drl_2}, and hybrid SLAM-based approaches~\cite{multi_drl}, train robots via extensive offline interactions. However, these methods often suffer from inefficient exploration due to limited semantic guidance and poor generalizability, as they require vast amounts of training data. In contrast, our open-vocabulary value map leverages foundation model cues to enhance semantic understanding and improve generalizability.

\subsection{Probabilistic Planning Methods for MOS}
Probabilistic planning methods, typically formulated as POMDPs~\cite{mos_base,mos_new,mos_system}, manage uncertainty in object locations and sensor noise by maintaining belief states. While effective at handling observation uncertainty, these methods suffer from high computational complexity in large, cluttered environments and generally require environmental priors, limiting their adaptability to novel settings. Our method alleviates these issues by leveraging foundation model cues to simplify belief updates and reduce the action space, and by incorporating frontier-based exploration into the POMDP reward function to facilitate navigation in unfamiliar environments, thus enhancing both efficiency and robustness.

\subsection{Foundation Model--Guided Methods for MOS}
 Foundation model--guided approaches utilize both natural language and visual inputs to generate rich semantic priors for navigation. By leveraging the contextual understanding of foundation models---which encompass both LLMs~\cite{esc,cat,lvmn,zson} and VLMs~\cite{vlfm,opennav}---these methods inform target localization with improved semantic cues. Early work in SOS demonstrated that combining frontier mapping with such cues leads to efficient target localization~\cite{vlfm}. For MOS, the Finder~\cite{finder} was the first to incorporate foundation model cues by constructing a multi-layer value map to guide search, achieving promising results. However, Finder~\cite{finder} struggles to recover from detector noise and occlusions because it always navigates to frontiers (i.e., the boundaries between explored and unexplored regions), which limits its ability to revisit and correct for missed detections. Our framework integrates foundation model cues with frontier-based exploration and POMDP planning to overcome these challenges, enabling robust recovery from missed detections in complex, partially observed environments.

%% file: Methodology.tex
\section{Methodology}
\label{sec:methodology}

\subsection{Problem Formulation}
The MOS problem requires a mobile robot to search for a set of \( K \) static target objects in an unknown environment. The robot's state at time \( t \) is \( \mathbf{x}_r(t) = (x, y, \phi) \in \mathbb{R}^3 \), where \( (x, y) \) is its position and \( \phi \) its orientation. The environment contains \( L \) static objects \( \mathcal{O}_{\text{env}} = \{ o_{s1}, o_{s2}, \dots, o_{sL} \} \), among which the target objects \( \mathcal{O}_{\text{tgt}} = \{ o_{t1}, \dots, o_{tK} \} \subseteq \mathcal{O}_{\text{env}} \) are to be located at unknown positions \( \mathbf{x}_{tj} \). The objective of MOS is to locate all objects in \( \mathcal{O}_{\text{tgt}} \) while minimizing the cumulative travel distance \( d = \int_0^T \| \dot{\mathbf{x}}_r(t) \| dt \), where \( T \) is the total search time.

\subsection{Mapping}

\subsubsection{Multi-Layer Value Map and Object Map}
Following previous works  VLFM~\cite{vlfm} and Finder~\cite{finder}, we employ a pre-trained BLIP-2~\cite{blip2} vision-language model to compute cosine similarity scores between the robot’s current RGB observation and text prompts corresponding to each target object. At each time step, a cone-shaped confidence mask is generated to represent the camera’s field of view (FOV). The confidence of each pixel is computed as:

\begin{equation}
c(i, j) = \cos^2 \left(\frac{\theta}{\theta_{\text{FOV}} / 2} \times \frac{\pi}{2} \right)
\end{equation}

where \( \theta \) is the angle between the pixel and the optical axis, and \( \theta_{\text{FOV}} \) is the camera's horizontal field of view.

To handle overlapping observations over time, we apply a weighted averaging update for the value map:

\begin{equation}
v_{i,j}^{\text{new}} = \frac{c_{i,j}^{\text{curr}} v_{i,j}^{\text{curr}} + c_{i,j}^{\text{prev}} v_{i,j}^{\text{prev}}}{c_{i,j}^{\text{curr}} + c_{i,j}^{\text{prev}}}
\end{equation}

where \( v_{i,j} \) represents the value at pixel \( (i,j) \), and \( c_{i,j} \) denotes the confidence score. The confidence update is computed as:

\begin{equation}
c_{i,j}^{\text{new}} = \frac{(c_{i,j}^{\text{curr}})^2 + (c_{i,j}^{\text{prev}})^2}{c_{i,j}^{\text{curr}} + c_{i,j}^{\text{prev}}}
\end{equation}

which biases the update towards higher confidence values.

After obtaining the value maps for different target objects, we normalize and aggregate them to form a shared value representation. This step follows a similar approach to previous works~\cite{mos_base,mos_new,mos_system} that assume independence across objects, where the joint belief is computed as the product of individual beliefs.

For object detection, we incorporate Grounding DINO~\cite{dino}, YOLOv7~\cite{yolo}, and Segment Anything Model (SAM)~\cite{sam}. These models enable us to detect, segment, and store all identified objects in an object map throughout the search process.

\subsubsection{Obstacle Map and Frontier Map}
We utilize depth and odometry data to construct a top-down 2D obstacle map, representing regions that the robot has identified as non-traversable. Based on this obstacle map, we determine boundaries between explored and unexplored areas and extract midpoints along these boundaries as potential frontier waypoints. These frontiers guide exploration in unknown environments.

\begin{algorithm}
\caption{\textbf{POUCT-based Planning} $(\mathcal{P}, b_t, d) \to {a}$}
\label{alg:vlm-pouct}
\begin{algorithmic}[1]
\Require $\mathcal{P} = \langle {\mathcal{S}}, {\mathcal{A}}, {\mathcal{O}}, {T}, {O}, R,\gamma \rangle$ 
\Statex \hspace{1.2cm} where
\Statex \hspace{1.2cm}  ${\mathcal{A}} = MoveTo\big\{(x_1^c, y_1^c), \dots, (x_n^c, y_n^c),(x^f, y^f)\big\}$ 
\Statex \hspace{1.2cm} $ b_t = \big\{( (x_1^c, y_1^c): p_1 ), \dots, \big( (x_n^c, y_n^c): p_n \big) \big\}$ 
\Ensure ${a}$: An action in the $\mathcal{A}$ of $\mathcal{P}$
\vspace{0.5em}
\Procedure{Plan}{$b_t$}
    \State $\mathcal{G} \gets$ GenerativeFunction($\mathcal{P}$)
    \State $Q(b_t, {a}) \gets$ POUCT($\mathcal{G}, h_t$)
    \State \Return ${a}$
\EndProcedure
\end{algorithmic}
\end{algorithm}

\subsection{Planning}

We model the planning process within an Object-Oriented POMDP (OO-POMDP) framework~\cite{mos_base}. In our formulation, the state and observation spaces are decomposed with respect to a single \textit{target object}, and multi-object search is achieved via our multi-layer value map using \textit{add} and \textit{norm} operations. Our approach introduces two key modifications: a novel action space formulation and an alternative belief update mechanism for real-world execution.

\subsubsection{Update the \textit{Action Space} and \textit{Belief} with Candidate Points and Selected Frontier}

Let the raw value map be denoted as \( v(x,y) \) over the spatial domain. First, the frontier with the highest value is selected from \( v(x,y) \) as a representative exploratory point \((x^f, y^f)\). Then, to refine the value distribution for targeted search, we apply a decay function:
\begin{equation}
v'(x,y) = \frac{1}{1 + \exp\left(\frac{u(x,y) - \tau}{\kappa}\right)},
\label{eq:decay}
\end{equation}

where \( u(x,y) \) is the update count at location \((x,y)\), and \(\tau\) and \(\kappa\) are constants controlling the decay rate. After thresholding \( v'(x,y) \) to extract high-value regions, we employ DBSCAN~\cite{dbscan} clustering to yield a set of candidate points $(x_i^c, y_i^c)$:
\[
\mathcal{C} = \{(x_i^c, y_i^c) \mid i = 1,\dots,n\}.
\]

The action space is then defined as:
\[
\mathcal{A} = \mathcal{A}_{\text{cand}} \cup \mathcal{A}_{\text{frontier}},
\]
with
\[
\mathcal{A}_{\text{cand}} = \{ \texttt{MoveTo}((x_i^c, y_i^c)) \mid i = 1,\dots,n \},
\]
\[
\mathcal{A}_{\text{frontier}} = \{ \texttt{MoveTo}((x^f, y^f)) \}.
\]
This formulation balances targeted search (via candidate points) with exploratory actions (via the frontier).

The belief over the target object's location is represented as a discrete distribution over candidate points:
\[
b_t = \{\, ((x_i^c, y_i^c): p_i) \mid i = 1,\dots,n \,\},
\]
where
\[
p_i = \frac{v(x_i^c, y_i^c)}{\sum_{j=1}^{n} v'(x_j^c, y_j^c)}.
\]
Here, \(p_i\) denotes the probability associated with the candidate point \((x_i^c, y_i^c)\).

During the POMDP solution process (i.e., in the simulation phase), we still update the belief using the standard Bayesian rule:
\[
b_{t+1}(s') = \eta\, \Pr(o \mid s', a) \sum_{s \in \mathcal{S}} \Pr(s' \mid s, a)\, b_t(s),
\]
with normalization constant \(\eta\). However, in real-world execution, after receiving an observation, we approximate the belief update using a \emph{decayed value map}. This approach circumvents explicit reliance on the observation model while retaining adaptability, and thereby offers a more efficient method for adjusting the belief in practical settings.

The rationale for this approach is that the value map itself already encodes a highly reliable representation of the probability that the object is located at each candidate point. After incorporating detector information into the decayed value map, there is no longer a need to perform extensive POMDP simulations—where possible observations are hypothesized and used in Bayesian updates—to adjust the belief after receiving the real observation. Instead, the decayed value map directly reflects the updated likelihood of the object's location in a simpler, yet still effective, manner.

\subsubsection{POMDP Components}

We model our planning problem as a POMDP defined by the tuple
\[
\langle \mathcal{S}, \mathcal{A}, \mathcal{O}, T, O, R, \gamma \rangle.
\]
The components are defined as follows:

\begin{itemize}
    \item \textbf{State Space (\(\mathcal{S}\))}: An environment state \( s \) is represented as a combination of the robot's state \( s_r \) and the target object's state \( s_t \):
    \[
    s = \{ s_r, s_t \}.
    \]
    The robot state is defined as \( s_r = (x_r, y_r) \) with \( (x_r, y_r) \in \mathbb{R}^2 \), representing its 2D position, while the target object's state is given by \( s_t = (x_t, y_t) \) with \( (x_t, y_t) \in \mathbb{R}^2 \), indicating its position in the environment.

    \item \textbf{Observation Space (\(\mathcal{O}\))}: The robot obtains an observation \( o \) from its RGB-D camera, modeled as a binary indicator of whether detected or not:
    \[
    o \in \{0,1\}.
    \]
    \item \textbf{Action Space (\(\mathcal{A}\))}: Actions are defined as movement commands toward a goal location:
    \[
    a = \texttt{MoveTo}(g)
    \]
    where \( g \) is chosen from candidate points extracted from the value map or the selected frontier.
     \item \textbf{Transition Model (\(T(s,a,s')\))}: The target object is static (i.e., \( s_t' = s_t \)), and the robot state transitions deterministically:
    \[
    s_r' = f(s_r, a),
    \]
    where \( f \) updates the robot's 2D position based on the selected action.
     \item \textbf{Observation Function  (\(O\))}:
    The detection probability is defined as:
   \begin{equation}
    \Pr(o=1 \mid s',a) =
    \begin{cases}
      1, \text{if } d(s_r', s_t) \leq \delta, \\[2mm]
      \exp\Bigl(-\beta\,\bigl(d(s_r', s_t)-\delta\bigr)\Bigr), \\[1mm]
      \quad \text{if } d(s_r', s_t) > \delta.
    \end{cases}
    \end{equation}

    where \( d(s_r', s_t) \) is the Euclidean distance between the robot and the target, \( \delta \) is the detection threshold, and \( \beta > 0 \) controls the decay rate.

    \

    \item \textbf{Reward Function (\(R(s,a)\))}: The reward function is defined as:
    \begin{equation}
    \begin{split}
    R(s,a) =\ & -\lambda_{\text{move}}\, d(s_r, s_r') \\
              & + \lambda_{\text{frontier}}\, \mathbb{I}(a \in \mathcal{A}_{\text{frontier}}) \\
              & + \lambda_{\text{target}}\, \mathbb{I}\Bigl(d(s_r', s_t) \leq \delta\Bigr),
    \end{split}
    \end{equation}
    where \( d(s_r, s_r') \) is the distance traveled by the robot, \(\mathbb{I}(\cdot)\) is the indicator function, and \(\lambda_{\text{move}}, \lambda_{\text{frontier}}, \lambda_{\text{target}} \ge 0\) are weight parameters balancing movement cost, exploratory incentive, and target proximity reward.

    \item \textbf{Discount Factor (\(\gamma\))}: A constant \(\gamma \in (0,1)\) is used to balance immediate and future rewards.
\end{itemize}

To solve the formulated POMDP efficiently, we employ the partially Observable Upper Confidence Trees (POUCT) algorithm~\cite{pouct}—a Monte Carlo Tree Search-based method. As illustrated in Algorithm~\ref{alg:vlm-pouct}, after the agent executes a step in the real environment, it obtains an updated map from which new candidate points and a frontier are extracted. These are then used to update the action space \( \mathcal{A} \) and the belief \( b_t \). 

Then, it is straightforward to define a generative function
\[
\mathcal{G}(s, a) \rightarrow (s', o, r),
\]
using its transition, observation, and reward functions. Leveraging this generative function, POUCT builds a search tree to simulate transitions, observations, and rewards, thereby planning and selecting the next optimal action.

%% file: results.tex
\section{Experiment}
\label{sec:results}
\begin{figure*}[t!]
    \centering
    \includegraphics[width = \textwidth]{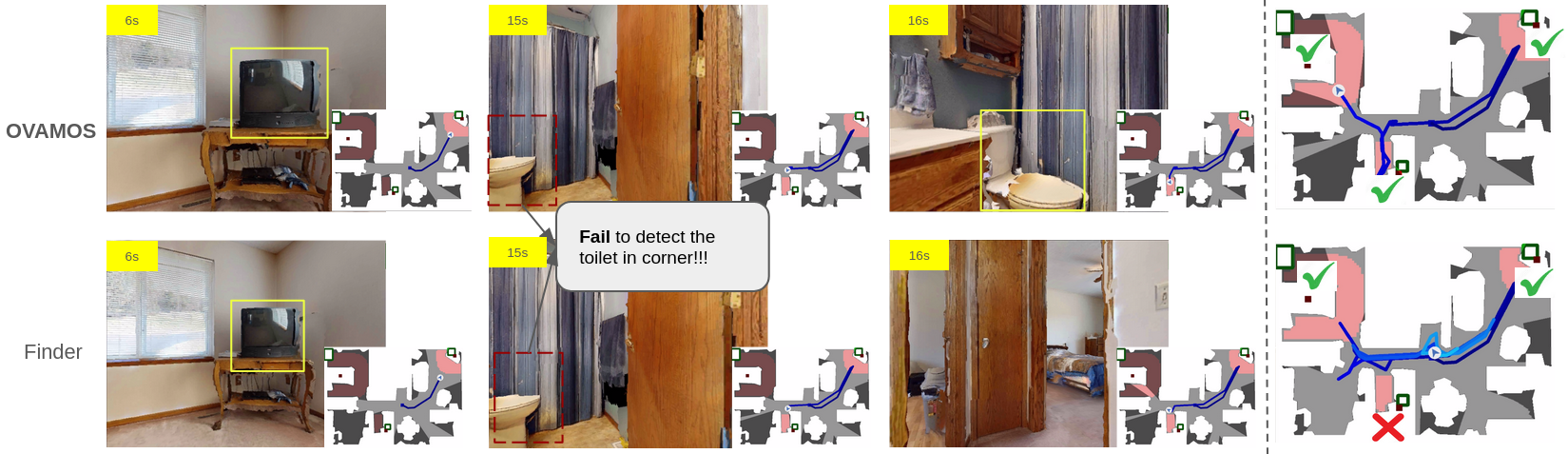}
    \caption{\textbf{Qualitative comparison between OVAMOS and Finder~\cite{finder} in simulation.}.}
    \label{fig:sim_qua}
\end{figure*}

\label{sec:results}
\begin{figure*}[t!]
    \centering
    \includegraphics[width = \textwidth]{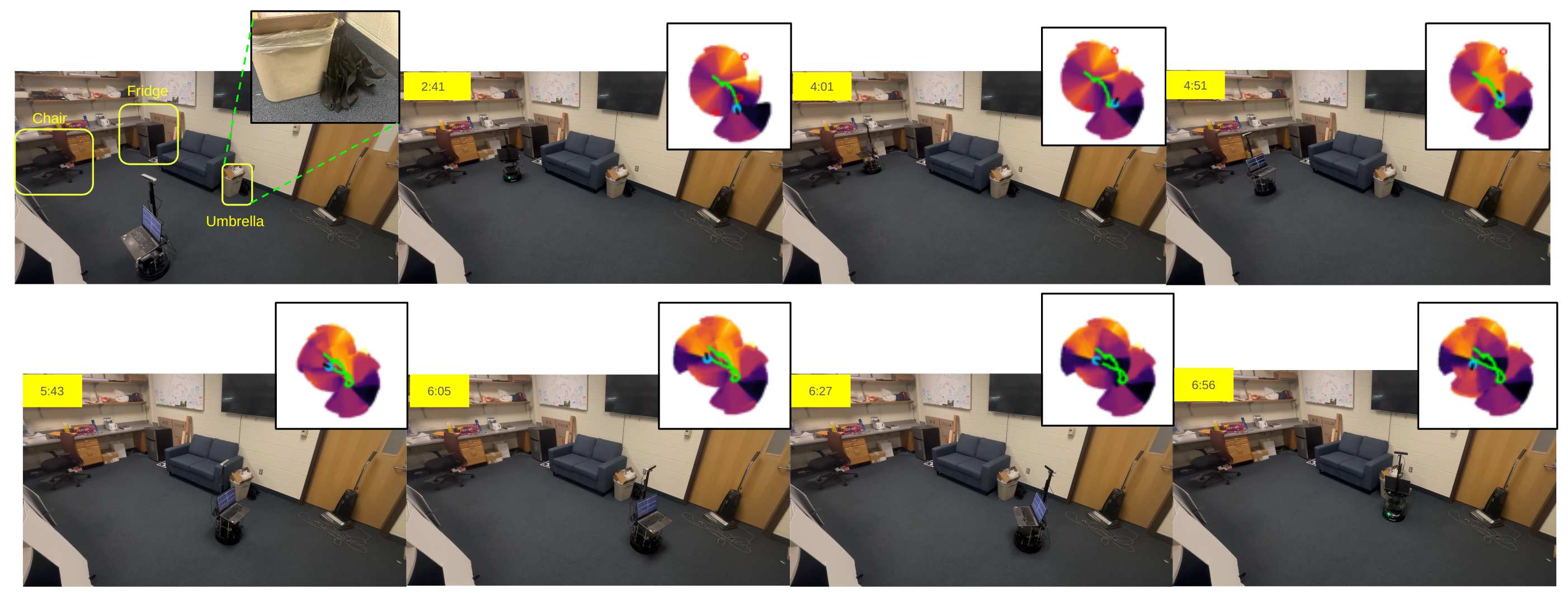}
    \caption{\textbf{Qualitative result of OVAMOS in real-world experiment.} }

    \label{fig:real_qual}
\end{figure*}

\begin{table}[t]
    \centering
    \setlength{\tabcolsep}{6pt} 
    \begin{tabular}{l|cc}
        \hline
        Methods & SR↑ & MSPL↑ \\
        \hline
        Random Walk & 0.0\%  & 0.0  \\
        VLFM    & 12.5\%  & 0.075  \\
        Finder  & 28.3\%  & 0.198  \\
        OVAMOS(Ours)    & \textbf{55.0\%}  & \textbf{0.497}  \\
        \hline
    \end{tabular}
    \caption{Overall performance comparison of Random Walk, VLFM~\cite{vlfm}, Finder~\cite{finder}, and Our Method on MOS, including Success Rate (SR) and Multi-Object Success weighted by normalized inverse Path Length (MSPL).}
    \label{tab:multi_object_search}
\end{table}

\begin{table}[t]
    \centering
    \setlength{\tabcolsep}{4.5pt} 
    \begin{tabular}{l|cc}
        \hline
        Methods & SR↑ & MSPL↑ \\
        \hline
        OVAMOS w/o POMDP & 27.5\%  & 0.154  \\
        OVAMOS w/o value decay & 45.0\%  & 0.410 \\
        OVAMOS (Full) & \textbf{55.0\%}  & \textbf{0.497}  \\
        \hline
    \end{tabular}
    \caption{Ablation study on OVAMOS, evaluating the impact of different components, including POMDP and value decay introduced in \ref{eq:decay}. 
    }
    \label{tab:ablation_study}
\end{table}

Our experiments consist of two stages: (1) a quantitative evaluation in a simulation environment, where we compare the success rate and efficiency of our method against recent state-of-the-art (SOTA) approaches and conduct an ablation study to analyze the impact of different components in OVAMOS, and (2) a qualitative validation on a real robot, where we deploy the agent in a \( 50 \, m^2 \) office environment to search for multiple objects.

\subsection{Simulation Experiments}

\subsubsection{Setup}  
We conduct our simulation experiments using the HM3D dataset's validation split. We select five scenes and construct a total of 120 episodes, all of which involve searching for \textit{three objects}. At the beginning of each episode, the robot is initialized at a random location within the environment and provided with a list of target objects. The episode progresses as follows: each time the robot calls \textit{stop}, it indicates that it has found an object. If the distance between the robot and the nearest target object is less than 1\,m at this moment, the object is considered successfully found. The robot then proceeds to search for the remaining objects. The episode terminates when either all target objects have been found or the robot reaches the maximum step limit of 500.

\subsubsection{Experimental Parameter Settings}
We set the following key parameters in our experiments: the reward function weights are set to \(\lambda_{\text{move}} = 20\), \(\lambda_{\text{frontier}} = 500\), and \(\lambda_{\text{target}} = 500\), which balance the cost of movement, the incentive for exploring frontiers, and the reward for proximity to the target, respectively. The discount factor is fixed at \(\gamma = 0.95\). In the observation function, the detection threshold is defined as \(\delta = 1\) m and the decay parameter \(\beta=1\) controls the exponential decay of detection probability for distances greater than \(\delta\).

\subsubsection{Evaluation Metrics}
To evaluate performance, we use three key metrics:

\begin{itemize}
    \item \textbf{Success Rate (SR)}: The percentage of episodes in which the robot successfully finds all target objects.
    \item \textbf{Multi-Object Success weighted by normalized inverse Path Length (MSPL)}: Based on the SPL metric, MSPL is calculated as:
    \begin{equation}
        MSPL = \frac{1}{N} \sum_{i=1}^{N} S_i \frac{l_i}{\max(p_i, l_i)}
    \end{equation}
    where $N$ denotes the total number of episodes, $S_i$ is a binary indicator of success for episode $i$, $l_i$ represents the optimal shortest path length from the start location to all target objects, and $p_i$ denotes the actual path length traversed by the robot.
\end{itemize}

\subsubsection{Comparison Baselines}
We compare our approach, OVAMOS, against the following baseline methods:
\begin{itemize}
    \item \textbf{Random Walk (Lower Bound)}: A naive strategy where the robot moves randomly in the environment.
    \item \textbf{VLFM~\cite{vlfm}(SOTA for SOS)}: This method combines a VLM-generated value map with frontier-based exploration, selecting the best frontier at each step. In our experiments, the MOS task is decomposed into a sequence of SOS tasks executed independently.
    \item \textbf{Finder~\cite{finder} (SOTA for MOS)}: Built upon VLFM, this method utilizes value maps to compute the \textit{scene-to-object} score and incorporates an additional \textit{object-to-object} score to select the optimal frontier.
\end{itemize}

\subsubsection{Results and Analysis}  
The experimental results are summarized in Table~\ref{tab:multi_object_search}. VLFM achieves only a 12.5\% success rate (SR) with an MSPL of 0.075, which reflects its limitation of tracking only one object at a time. Finder improves on this by employing a multi-layer value map to simultaneously track multiple objects, resulting in an enhanced SR of 28.3\% and an MSPL of 0.198. Our method, OVAMOS, further builds on Finder’s strengths by integrating POMDP-based planning into the framework. This additional planning mechanism enables adaptive exploration and more informed decision-making, leading to a significant performance boost with an SR of 55.0\% and an MSPL of 0.497. Thus, OVAMOS nearly doubles the success rate and more than doubles the MSPL compared to Finder, demonstrating its superior accuracy and efficiency in multi-object search tasks.

The key advantage of OVAMOS over Finder lies in scenarios where objects are occluded or partially visible. As illustrated in Fig.~\ref{fig:sim_qua}, in some episodes, due to occlusions and viewing angles, a toilet located in the corner was not detected in the already-explored region. Finder, relying on its frontier-based strategy, abandoned this area and navigated toward a different frontier, ultimately never revisiting the corner and failing to find the toilet. In contrast, OVAMOS recognized the region's high value and continued further exploration, eventually obtaining an effective viewpoint to detect the object. This ability to revisit and refine exploration based on value map information provides OVAMOS with a distinct advantage over Finder in complex search scenarios.We did not compare with VLFM and Random Walk here because their performance is significantly weaker than OVAMOS and Finder.

In addition, we conduct an ablation study to analyze the impact of key components in OVAMOS, specifically the absence of POMDP-based planning and the absence of value decay. The results in Table~\ref{tab:ablation_study} indicate that removing POMDP leads to much worse performance. Without POMDP, the agent directly selects the highest-value frontier or candidate point at each step, resulting in an overly greedy strategy. This approach fails to account for the navigation effort required to reach each point and does not balance the trade-off between exploring new frontiers and exploiting high-value candidate points, leading to suboptimal overall performance.

For the case without value decay in TABLE~\ref{tab:ablation_study}, we observe a drop in success rate. Further analysis reveals that in some episodes, certain regions maintained high-value scores despite not containing the actual target objects. As a result, the robot spent excessive steps repeatedly exploring these high-value areas without making progress, sometimes getting trapped and ultimately failing the task. This highlights the importance of value decay in preventing the robot from fixating on misleading high-value regions and ensuring more effective search behavior.

\subsection{Physical Experiments}
\subsubsection{Setup}
As shown in Fig.~\ref{fig:intro} and Fig.~\ref{fig:real_qual}, we deployed OVAMOS on a TurtleBot platform. All algorithms were executed on a laptop equipped with an 8\,GB GPU. Due to GPU memory constraints, we replaced BLIP2~\cite{blip2} from the previous framework with BLIP~\cite{blip}. The robot was then tasked with searching for objects in a cluttered \(50\,m^2\) office environment, including easily detectable objects (a chair and a fridge) and a more challenging object (an umbrella occluded by a trash can). 

\subsubsection{Results}
The complete task was executed in 6'56". As shown in Fig.~\ref{fig:real_qual}, initially the fridge and chair were easily detected due to their large size and clear features. In particular, the fridge was found at 2'41" and the chair at 4'01", while the umbrella remained hidden because of unfavorable viewpoints. Once the entire environment had been explored—and from 5'43" onward the value map no longer showed any frontiers—the robot continuously replanned its trajectory based on the updated value map information. This iterative replanning eventually led to an effective viewpoint from which the occluded umbrella was detected. 

This experiment demonstrates the strong recovery and replanning capability of OVAMOS in real-world scenarios, especially when the environment has been fully explored yet the target object remains undetected.

%% file: conclusion.tex
\section{Conclusion}
\label{sec:conclusion}
In this work, we integrate VLMs, frontier-based exploration, and POMDP to construct a more robust and efficient multi-object search framework compared to previous approaches~\cite{finder,vlfm}. We have validated the robustness of our algorithm in both simulation and real-world experiments. Extensive evaluations show that compared to the state-of-the-art, our method achieves a success rate improvement of 26.7\% and an MSPL improvement of 0.3, highlighting its superior performance in both effectiveness and efficiency.

However, our approach has certain limitations. Currently, our framework is constrained to 2D environments, the robot's action space is relatively simple, and the camera viewpoint remains fixed. These constraints make our method insufficient for tackling the complexity of real-world MOS tasks.

As future work, we aim to extend the search space to 3D environments, incorporate more complex actions such as moving occlusions, and enable dynamic camera adjustments to obtain better viewpoints. These enhancements will further improve the adaptability and effectiveness of our approach.